\documentclass[runningheads]{llncs}
\usepackage{graphicx}
\usepackage[english]{babel}
\usepackage[utf8]{inputenc}
\usepackage[caption=false,font=footnotesize]{subfig}
\usepackage[export]{adjustbox}
\usepackage[labelfont=bf]{caption}
\usepackage{booktabs}
\usepackage{amsmath,amssymb}
\usepackage[pagebackref=true]{hyperref}

\usepackage[table]{xcolor}
\usepackage{collcell}
\usepackage{hhline}
\usepackage{pgf}
\usepackage{multirow}
\usepackage{hyperref}
\usepackage{blindtext}
\usepackage{enumitem}
\usepackage{bbm}
\usepackage{siunitx}
\usepackage{arydshln}

\usepackage{accents}

\usepackage{wrapfig}
\usepackage{bm}
\usepackage{soul}
\usepackage[normalem]{ulem}
\usepackage{makecell}
\def\colorModel{hsb} %
\newcommand\ColCell[1]{
  \pgfmathparse{#1<50?1:0}  %
    \ifnum\pgfmathresult=0\relax\color{white}\fi
  \pgfmathsetmacro\compA{0}      %
  \pgfmathsetmacro\compB{#1/100} %
  \pgfmathsetmacro\compC{1}      %
  \edef\x{\noexpand\centering\noexpand\cellcolor[\colorModel]{\compA,\compB,\compC}}\x #1
  } 
\newcolumntype{E}{>{\collectcell\ColCell}m{0.45cm}<{\endcollectcell}}  %

\usepackage[T1]{fontenc} 
\usepackage{lipsum}
\usepackage{tcolorbox}

\newtcolorbox{afancybox}[1][]{#1,colback=white,colframe=black}
\usepackage{ulem}

\usepackage{commath}
\usepackage[misc,geometry]{ifsym}
\usepackage[title]{appendix}
\usepackage[figuresright]{rotating}
\usepackage{booktabs}
\usepackage{xfrac}
\usepackage{svg}

\newcommand{\y}{\mathbf{y}}
\newcommand{\p}{\mathbf{p}}
\newcommand{\rps}{\textrm{RPS}}

\newcommand{\unl}[1]{\underline{#1}}
\newcommand{\ft}[1]{\underline{\textbf{#1}}}
\newcommand{\sd}[1]{\textbf{#1}}
\newcommand{\xdownarrow}[1]{%
  {\left\downarrow\vbox to #1{}\right.\kern-\nulldelimiterspace}
}

\begin{document}

\title{Performance Metrics for \\Probabilistic Ordinal Classifiers}

\author{Adrian Galdran\inst{1,2}$^\textrm{,\Letter}$}
\authorrunning{A. Galdran}
\institute{BCN Medtech, Universitat Pompeu Fabra, Barcelona, Spain, \email{adrian.galdran@upf.edu}
\and
University of Adelaide, Adelaide, Australia}
\maketitle              %
\begin{abstract}
Ordinal classification models assign higher penalties to predictions further away from the true class. 
As a result, they are appropriate for relevant diagnostic tasks like disease progression prediction or medical image grading. 
The consensus for assessing their categorical predictions dictates the use of distance-sensitive metrics like the Quadratic-Weighted Kappa score or the Expected Cost. 
However, there has been little discussion regarding how to measure performance of probabilistic predictions for ordinal classifiers. 
In conventional classification, common measures for probabilistic predictions are Proper Scoring Rules (PSR) like the Brier score, or Calibration Errors like the ECE, yet these are not optimal choices for ordinal classification. 
A PSR named Ranked Probability Score (RPS), widely popular in the forecasting field, is more suitable for this task, but it has received no attention in the image analysis community. 
This paper advocates the use of the RPS for image grading tasks. 
In addition, we demonstrate a counter-intuitive and questionable behavior of this score, and propose a simple fix for it. 
Comprehensive experiments on four large-scale biomedical image grading problems over three different datasets show that the RPS is a more suitable performance metric for probabilistic ordinal predictions. 
Code to reproduce our experiments can be found at \url{github.com/agaldran/prob_ord_metrics}.
\end{abstract}

\setcounter{footnote}{0} 

\section{Introduction and Related Work}
The output of predictive machine learning models is often presented as categorical values, \textit{i.e.} ``hard'' class membership decisions. 
Nonetheless, understanding the faithfulness of the underlying probabilistic predictions giving rise to such hard class decisions can be essential in some critical applications. 
Meaningful probabilities enable not only high model accuracy, but also more reliable decisions: a doctor may choose to order further diagnostic tests if a binary classifier gives a $p=45\%$ probability of disease, even if the hard prediction is ``healthy'' \cite{cahan_probabilistic_2003}. 
This is particularly true for ordinal classification problems, \textit{e.g.} disease severity staging \cite{galdran_non-uniform_2020,galdran_cost-sensitive_2020} or medical image grading \cite{jaroensri_deep_2022,silva-rodriguez_going_2020}. 
In these problems, predictions should be \textit{as close as possible to the actual category}; further away predictions must incur in heavier penalties, as they have increasingly worse consequences. 

There is a large body of research around performance metrics for medical image analysis \cite{reinke_understanding_2023}. 
Most existing measures, like accuracy or the F1-score, focus on assessing hard predictions in specific ways that capture different aspects of a problem. 
In ordinal classification, the recommended metrics are Quadratic-Weighted Kappa and the Expected Cost \cite{ferrer_analysis_2022,maier-hein_metrics_2023}.
In recent years, measuring the performance of ``soft'' probabilistic predictions has attracted an increasing research interest \cite{gruber_better_2022,perez-lebel_beyond_2023}. 
For this purpose, the current consensus is to employ Calibration Errors like the ECE and Proper Scoring Rules like the Brier score \cite{maier-hein_metrics_2023}. 
We will show that other metrics can instead be a better choice for assessing probabilistic predictions in the particular case of ordinal classification problems.

How to measure the correctness of probabilistic predictions is a decades-old question, naturally connected to forecasting, \textit{i.e.} predicting the future state of a complex system \cite{gneiting_probabilistic_2014}. 
A key aspect of forecasting is that, contrary to classifiers, forecasters do not output hard decisions, but probability distributions over possible outcomes. 
Weather forecasts do not tell us whether it will rain tomorrow or not, they give us a probability estimate about the likelihood of raining, leaving to us the decision of taking or not an umbrella, considering the personal cost of making such decision. 
The same applies for financial investments or sports betting, where it is also the final user who judges risks and makes decisions based on probabilistic forecasts. 
In this context, Proper Scoring Rules (PSRs) have been long used by the forecasting community to measure predictive performance \cite{gneiting_strictly_2007}.
PSRs are the focus of this paper, and will be formally defined in section \ref{a}.

\paragraph{Relation to Calibration:} A popular approach to assess the quality of probabilistic predictions is measuring calibration. 
A model is well calibrated if its probabilistic predictions are aligned with its accuracy on average.
PSRs and calibration are intertwined concepts: PSRs can be decomposed into a calibration and a resolution component \cite{gneiting_probabilistic_2007}. 
Therefore, a model needs to be both calibrated and resolved (\textit{i.e.} having \textit{sharp}, or \textit{concentrated} probabilities) in order to have a good PSR value. 
For example, if a disease appears in 60\% of the population, and our model is just ``\texttt{return p=0.6}'', in the long run the model is correct $60\%$ of the time, and it is perfectly calibrated, as its confidence is fully aligned with its accuracy, despite having zero predictive ability. 
If the model predicted in a ``resolved'' manner with $p=0.99$ the presence of the disease, but being correct only $70\%$ of the time, then it would be overconfident, which is a form of miscalibration. 
Only when the model is simultaneously confident and correct can it attain a good PSR value.

The two most widely adopted PSRs are the Brier and the Logarithmic Score \cite{brier_verification_1950,good_rational_1952}. 
Unfortunately, none of these is appropriate for the assessment of ordinal classification probabilities \cite{constantinou_solving_2012}. 
A third PSR, long used by forecasting researchers in this scenario, the Ranked Probability Score (RPS, \cite{epstein_scoring_1969}), appears to have been neglected so far in biomedical image grading applications. 
This paper first covers the definition and basic properties of PSRs, and then motivates the use the RPS for ordinal classifiers. 
We also illustrate a counter-intuitive behavior of the RPS, and propose a simple modification to solve it.
Our experiments cover two relevant biomedical image grading problems and illustrate how the RPS can better assess probabilistic predictions of ordinal classification models.

\section{Methods}

\subsection{Scoring Rules - Notation, Properties, Examples}\label{a}
We consider a $K$-class classification problem, and a classifier that takes an image $\mathbf{x}$ and maps it into a vector of probabilities $\p\in [0,1]^K$. 
Typically, $\p$ is the result of applying a softmax operation on the output of a neural network. 
Suppose $\mathbf{x}$ belongs to class $y\in\{1,...,K\}$, and denote by $\y$ its one-hot representation. 
A Scoring Rule (SR) $\mathcal{S}$ is any function taking the probabilistic prediction $\p$ and the label $\y$ and producing a number $\mathcal{S}(\p, \y)\in \mathbb{R}$ (a score). 
Here we consider negatively oriented SRs, which assign lower values to \textit{better predictions}. 

Of course, the above is an extremely generic definition, to which we must now attach additional properties in order to encode our understanding of what \textit{better predictions} means for a particular problem.

\paragraph{\textbf{Property 1:}} A Scoring Rule (SR) is \textit{proper} if its value is minimal when the probabilistic prediction coincides with the ground-truth in expectation.

\paragraph{\textbf{Example:}} The Brier Score \cite{brier_verification_1950} is defined as the sum of the squared differences between probabilities and labels:
\begin{equation}\label{brier}
\textrm{Brier}(\p,\y) = \|\p-\y\|_2^2 = \sum_{i=1}^K (p_i - y_i)^2.
\end{equation}
Since its value is always non-negative, and it decreases to 0 when $\p=\y$, we conclude that the Brier Score is indeed proper.

\paragraph{\textbf{Property 2:}} A Proper Scoring Rule (PSR) is \textit{local} if its value only depends on the probability assigned to the correct category.

\paragraph{\textbf{Example:}} The Brier Score is non-local, as its value depends on the probability placed by the model on all classes. 
The Logarithmic Score \cite{good_rational_1952}, given by:
\begin{equation}\label{log}
\mathcal{L}(\p,\y) = -\log(p_c)
\end{equation}
where $c$ is the correct category of $\mathbf{x}$, rewards the model by placing as much probability mass as possible in $c$, regardless of how the remaining probability is distributed. 
It is, therefore, a local PSR. 
The Logarithmic Score is also known, when taken on average over a dataset, as the Negative Log-Likelihood.

\paragraph{\textbf{Property 3:}} A PSR is \textit{sensitive to distance} if its value takes into account the order of the categories, in such a way that probability placed in categories further away from the correct class is more heavily penalized.

\paragraph{\textbf{Example:}} Both the Brier and the Logarithmic scores are insensitive to distance (shuffling $\p$ and $\y$ won't affect the score). 
Sensitivity to distance is essential for assessing ordinal classifiers. 
Below we define the Ranked Probability Score (RPS) \cite{epstein_scoring_1969,murphy_note_1971}, which has this property, and is therefore more suitable for our purposes.

\subsection{The Ranked Probability Score for Ordinal Classification}\label{c}
\begin{figure}[!t]
\centerline{\includegraphics[width=0.9\textwidth]{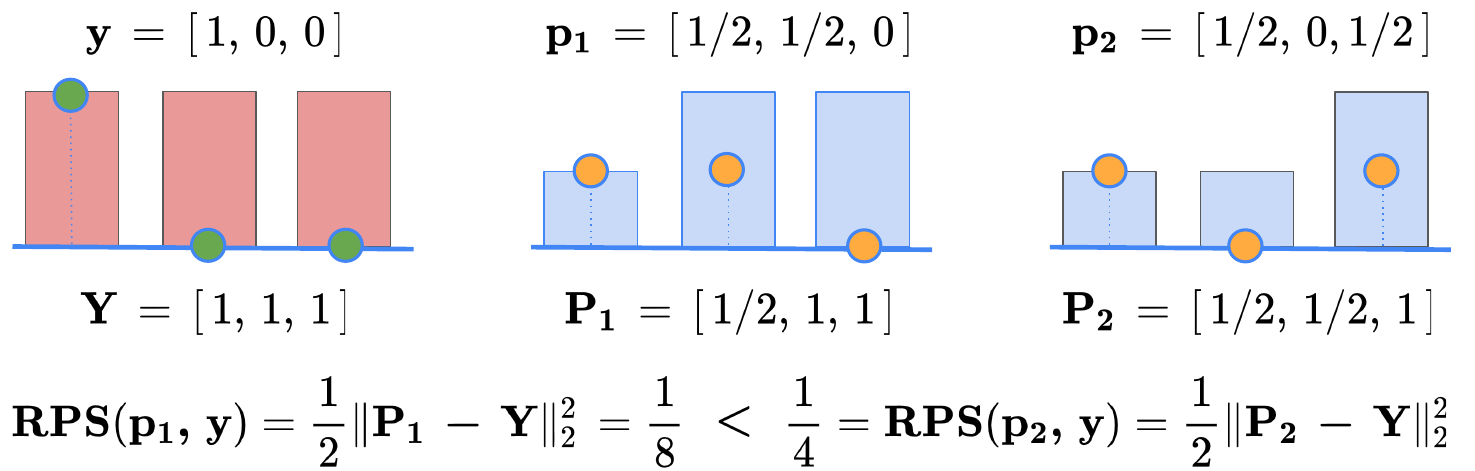}}
\caption{The RPS is sensitive to distance, suitable for assessing probabilistic predictions on biomedical image grading problems. 
It is the difference between the cumulative probability distributions of the label and a probabilistic prediction.}
\label{rps}
\end{figure}

Consider a test sample $(\mathbf{x},\y)$ in a 3-class classification problem, with label $\y$ and two probabilistic predictions $\p_1,\p_2$:
\begin{equation}\label{example}
\y=[\,1,0,0\,], \ \ \p_1=[\,\frac{1}{4},\frac{3}{4},0\,], \ \  \p_2=[\,\frac{1}{4},0,\frac{3}{4}\,]
\end{equation}
In this scenario, both the Brier and the Logarithmic scores produce the same penalty for each prediction, whereas a user might prefer $\p_1$ over $\p_2$ due to the latter assigning more probability to the second category. 
Indeed, if we use the arg-max operator to generate a hard-decision for this sample, we will obtain a prediction of class 2 and class 3 respectively, which could result in the second model declaring a patient as severely unhealthy with serious consequences. 
In this context, we would like to have a PSR that takes into account distance to the true category, such as the Ranked Probability Score (RPS, \cite{epstein_scoring_1969}), given by:
\begin{equation}\label{rps_eq}
\textrm{RPS}(\p, \y) = \frac{1}{K-1} \sum_{i=1}^{K-1} \left[\sum_{j=1}^i (p_j-y_j)\right]^2 = \frac{1}{K-1} \|\mathbf{P} - \mathbf{Y}\|^2_2.
\end{equation}
The RPS is the squared $\boldsymbol{\ell_2}$ distance between the cumulative distributions $\mathbf{Y}$ of the target label $\y$ and $\mathbf{P}$ of the probabilistic prediction $\p$, discounting their last component (as they are both always one) and normalizing so that it varies in the unit interval. 
In the above example, the RPS would give for each prediction a penalty of $\textrm{RPS}(\p_1, \y)=\sfrac{1}{8}$, $\textrm{RPS}(\p_2, \y)=\sfrac{1}{4}$, as shown in Figure \ref{rps}. 

Among many interesting properties, one can show that the RPS is proper \cite{murphy_ranked_1969}, and reduces to the Brier score for $K=2$. 
Despite the RPS dating back more than 50 years \cite{epstein_scoring_1969}, and enjoying great popularity in the weather forecasting community, it appears to be much less known in the image analysis and computer vision areas, where we could not find any trace of it. 
The \textbf{first goal} of this paper is to bring to the attention of computer vision researchers this tool for measuring the performance of probabilistic predictions in ordinal classification.

\begin{figure}[!b]
\caption{The Ranked Probability Score displays some counter-intuitive behavior that the proposed sa-RPS can fix. 
Here, $\p_2$ places more probability on the correct class but $\p_1$ is preferred due to its symmetry.}
\centerline{\includegraphics[width=1\textwidth]{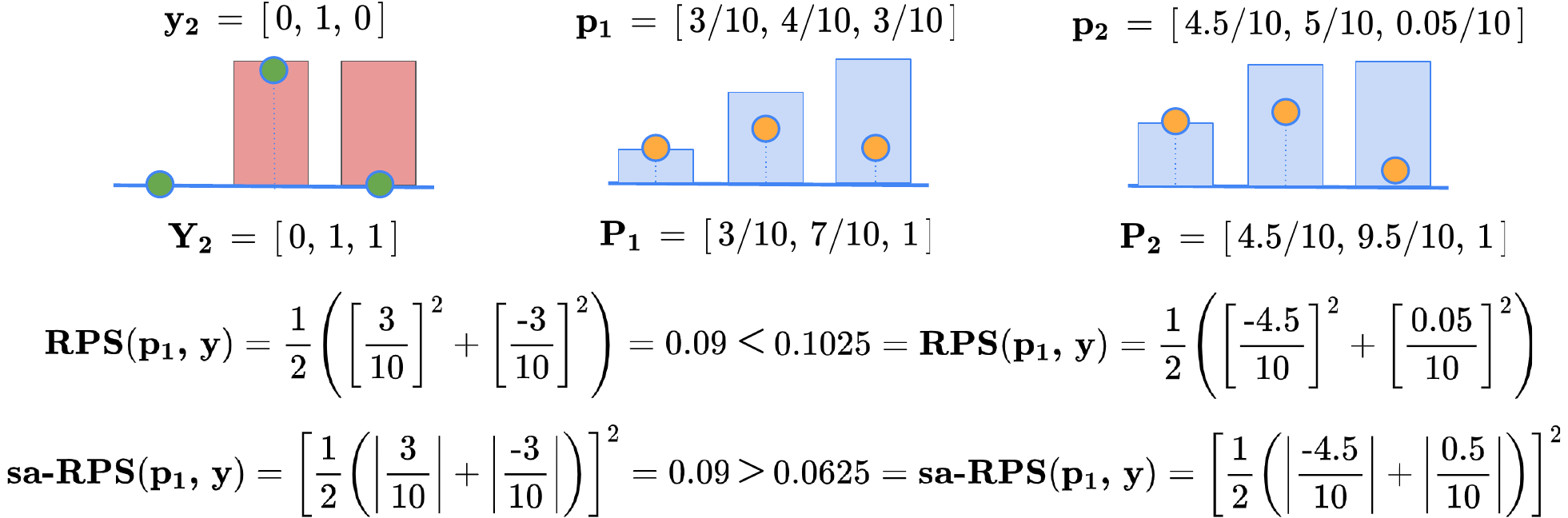}}
\label{sa_rps}
\end{figure}

\subsection{The Squared Absolute RPS}\label{d}
Our \textbf{second goal} in this paper is to identify and then fix certain failure modes of the RPS that might lead to counter-intuitive behaviors. 
First, in disease grading and other ordinal classification problems it is customary to assign penalties to mistakes that grow quadratically with the distance to the correct category. 
This is the reason why most works utilize the Quadratic-Weighted Kappa Score (QWK) instead of the linearly weighted version of this metric. 
However, the RPS increases the penalty linearly, as can be quickly seen with a simple 3-class problem and an example $(\mathbf{x}_1, \y_1)$ of class 1 ($\y_1=[\,1,0,0\,]$):
\begin{equation}\label{example1}
\rps([\,1,0,0\,], \y_1)=0, \ \ \rps([\,0,1,0\,], \y_1)=1/2. \ \ \rps([\,0,0,1\,], \y_1)=1.
\end{equation}
Also, the RPS has a hidden preference for symmetric predictions. 
To see this, consider a second example $(\mathbf{x}_2, \y_2)$ in which the correct category is now the middle one \textbf{($\y_2=[\,0,1,0\,]$)}, and two probabilistic predictions: $p_\mathrm{sym}=[\,3/10,4/10,3/10\,]$, $p_\mathrm{asym}=[\,1/10,5/10,9/10\,]$. 
In principle, there is no reason to prefer $p_\mathrm{sym}$ over $p_\mathrm{asym}$, unless certain prior/domain knowledge tells us that symmetry is a desirable property. 
In this particular case, $p_\mathrm{asym}$ is actually more confident on the correct class than $p_\mathrm{sym}$, which is however the preferred prediction for the RPS:
\begin{equation}\label{example2}
\rps([\,0.30,0.40,0.30\,], \y_2) = 0.09 < 0.1025 = \rps([\,0.45,0.50,0.05\,], \y_2).
\end{equation}
In order to address these aspects of the conventional RPS, we propose to implement instead the Squared Absolute RPS (sa-RPS), given by:
\begin{equation}
\textrm{sa-RPS}(\p, \y) = \frac{1}{K-1} \left[\sum_{i=1}^K \left|\sum_{j=1}^i (p_j-y_j)\right|\right]^2
\end{equation}
Replacing the inner square in eq. (\ref{rps_eq}) by an absolute value, we manage to break the preference for symmetry of the RPS, and squaring the overall result we build a metric that still varies in [0,1] but gives a quadratic penalty to further away predictions. 
This is illustrated in Fig. \ref{sa_rps} above.

\subsection{Evaluating Evaluation Metrics}\label{experiment_description}
Our \textbf{third goal} is to demonstrate how the (sa-)RPS is useful for evaluating probabilistic ordinal predictions. 
In the next section we will show some illustrative examples that qualitatively demonstrate its superiority over the Brier and logarithmic score. 
However, it is hard to quantitatively make the case for one performance metric over another, since metrics themselves are what quantify modeling success. 
We proceed as follows: we first train a neural network to solve a biomedical image grading problem. 
We generate probabilistic predictions on the test set and apply distance sensitive metrics to (arg-maxed) hard predictions (QWK and EC, as recommended in \cite{maier-hein_metrics_2023}), verifying model convergence. 

Here it is important to stress that, contrary to conventional metrics (like accuracy, QWK, or ECE) PSRs can act on an individual datum, without averaging over sets of samples.
We exploit this property to design the following experiment: 
we sort the probabilistic predictions of the test set according to a score $\mathcal{S}$, and then progressively remove samples that are of worst quality according to $\mathcal{S}$. 
We take the arg-max on the remaining probabilistic predictions and compute QWK and EC. 
If $\mathcal{S}$ prefers better ordinal predictions, we must see a performance increase on that subset. 
We repeat this process, each time removing more of the worse samples, and graph the evolution of QWK and EC for different scores $\mathcal{S}$: a better score should result in a faster QWK/EC-improving trend. 

Lastly, in order to derive a single number to measure performance, we compute the area under the remaining samples vs QWK/EC curve, which we call Area under the Retained Samples Curve (AURSC). 
In summary:

\begin{afancybox}[width=.99\textwidth]
\begin{center}
\uline{\textbf{What we expect to see:}}
\end{center}\vspace{-0.25cm}
As we remove test set samples considered as worse classified by RPS, we expect to more quickly improve QWK/EC on the resulting subsets.
We measure this with the Area under the Retained Samples Curve (AURSC)
\end{afancybox}

\section{Experimental Results}
We now give a description of the data we used for experimentation, analyze performance for each considered problem, and close with a discussion of results.

\subsection{Datasets and Architecture}\label{datasets}
Our experiments are on two different medical image grading tasks:
\textbf{1)} the \textbf{TMED}-v2 dataset (\cite{huang_tmed_2022}, \href{https://tmed.cs.tufts.edu/tmed_v2.html}{link}) contains 17,270 images from 577 patients, with an aortic stenosis (AS) diagnostic label from three categories (none, early AS, or significant AS).
The authors provide an official train/test distribution of the data that we use here.
\textbf{2)} \textbf{Eyepacs} (\href{https://www.kaggle.com/c/diabetic-retinopathy-detection}{link}) contains retinal images and labels for grading Diabetic Retinopathy (DR) stage into five categories, ranging from healthy to proliferative DR. 
Ithas 35,126 images for training and 53,576 in the test set.
  
We train a ConvNeXt \cite{liu_convnet_2022}, minimizing the CE loss with the adam algorithm for 10 epochs starting with a learning rate of $l=1e$-$4$, decaying to zero over the training. 
We report average Area under the Retained Samples Curve (AURSC) for $50$ bootstrap iterations in each dataset below, and also plot the evolution of performance as we remove more samples considered to be worse by four PSRs: the Brier score, the Logarithmic score (Neg-Log), RPS and sa-RPS. 

\begin{figure}[!b]
\caption{For the same test set and predictions, the RPS finds wrong samples that are more incorrect from the point of view of ordinal classification.}
\centerline{\includegraphics[width=1\textwidth]{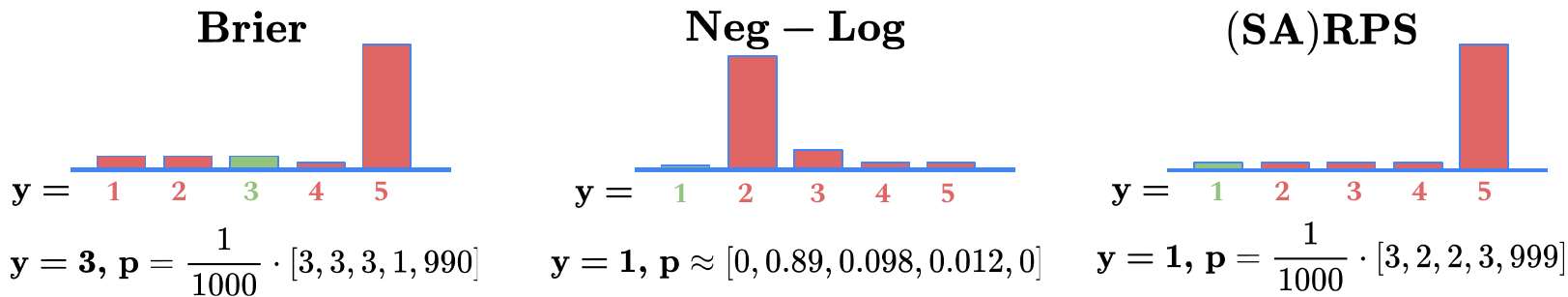}}
\label{fig_extra}
\end{figure}

\subsection{How is RPS useful? Qualitative Error Analysis}\label{qualitative}
The obvious application of RPS would be to train better ordinal classification models. 
But beyond this, RPS also enables improved, fine-grained error analysis. 
Let us see this through a simple experiment. 
Since PSRs assess samples individually, we can sort our test set using RPS, NLL, and Brier score.
The worst-scored items are what the model considers the wrongest probabilistic predictions.
The result of sorting predictions on the Eyepacs test set with the Brier, Neg-Log and RPS rules is show on Fig. \ref{fig_extra}. 
We can see that the prediction identified as worst by the RPS does indeed violate more heavily the order of categories, placing more probability on class 5 for a sample of class 1. 
On the other hand, for the same test set and predictions, the Brier score finds worst a prediction with 99\% of the probability on class 3 and a label of class 5, and the Neg-Log score identifies a sample of class 1 for which the model wrongly predicts class 2.

\subsection{Quantitative Experimental Analysis}\label{results}
Quantitative results of the experiment described in section \ref{experiment_description}, computing AURSC values for all PSRs, are shown in Table \ref{aursc_cvx_values}, with dispersion measures obtained from 50 bootstraped performance measurements. 

\begin{table}[!h]
\renewcommand{\arraystretch}{1.03}
\setlength\tabcolsep{3.50pt}
\begin{center}
\begin{tabular}{c cc cc}
& \multicolumn{2}{c}{\textbf{TMED}} &  \multicolumn{2}{c}{\textbf{Eyepacs}} \\
\cmidrule(lr){2-3} \cmidrule(lr){4-5} 
&  {\footnotesize\textbf{AURSC-QWK}$\uparrow$}  &  {\footnotesize\textbf{AURSC-EC}$\downarrow$}   &  {\footnotesize\textbf{AURSC-QWK}$\uparrow$}  &  {\footnotesize\textbf{AURSC-EC}$\downarrow$} \\
\midrule
\textbf{Brier}       & 13.46 $\pm$ 0.35      & 3.76 $\pm$ 0.21      & 17.36 $\pm$ 0.04      & 2.84 $\pm$ 0.07       \\
\midrule
\textbf{Neg-Log}     & 13.56 $\pm$ 0.35      & 3.62 $\pm$ 0.2       & 17.44 $\pm$ 0.04      & 2.67 $\pm$ 0.07       \\
\midrule
\textbf{RPS  }       & \sd{14.76 $\pm$ 0.28} & \sd{2.68 $\pm$ 0.14} & \sd{17.81 $\pm$ 0.03} & \sd{1.99 $\pm$ 0.04}       \\
\midrule
\textbf{sa-RPS}      & \ft{14.95 $\pm$ 0.25} & \ft{2.53 $\pm$ 0.12} & \ft{17.86 $\pm$ 0.03} & \ft{1.88$\pm$ 0.04}      \\
\bottomrule
\\[-0.25cm] 
\end{tabular}
\caption{Areas under the Retained Samples Curve for \textbf{TMED} and \textbf{Eyepacs}, with a \textbf{ConvNeXt}, for each PSR; 
\unl{\textbf{best}} and \textbf{second best} values are marked.}\label{aursc_cvx}
\end{center}
\vspace{-1cm}
\end{table}

We see that for the considered ordinal classification problems, dis\-tance-sensitive scores consistently outperform the Brier and Neg-Log scores.
Also, the Square-Absolute Ranked Probability Score always outperforms the conventional Ranked Probability Score. 
It is worth stressing that when observing bootstrapped performance intervals, neither the Brier nor the Logarithmic scores manage to overlap the SA-RPS interval in any of the two datasets, and in the Eyepacs dataset not even the best RPS result reaches the performance of worst SA-RPS result.

\begin{figure}[!b]
    \centering
    \subfloat{\includegraphics[width = 0.495\textwidth,valign=c]{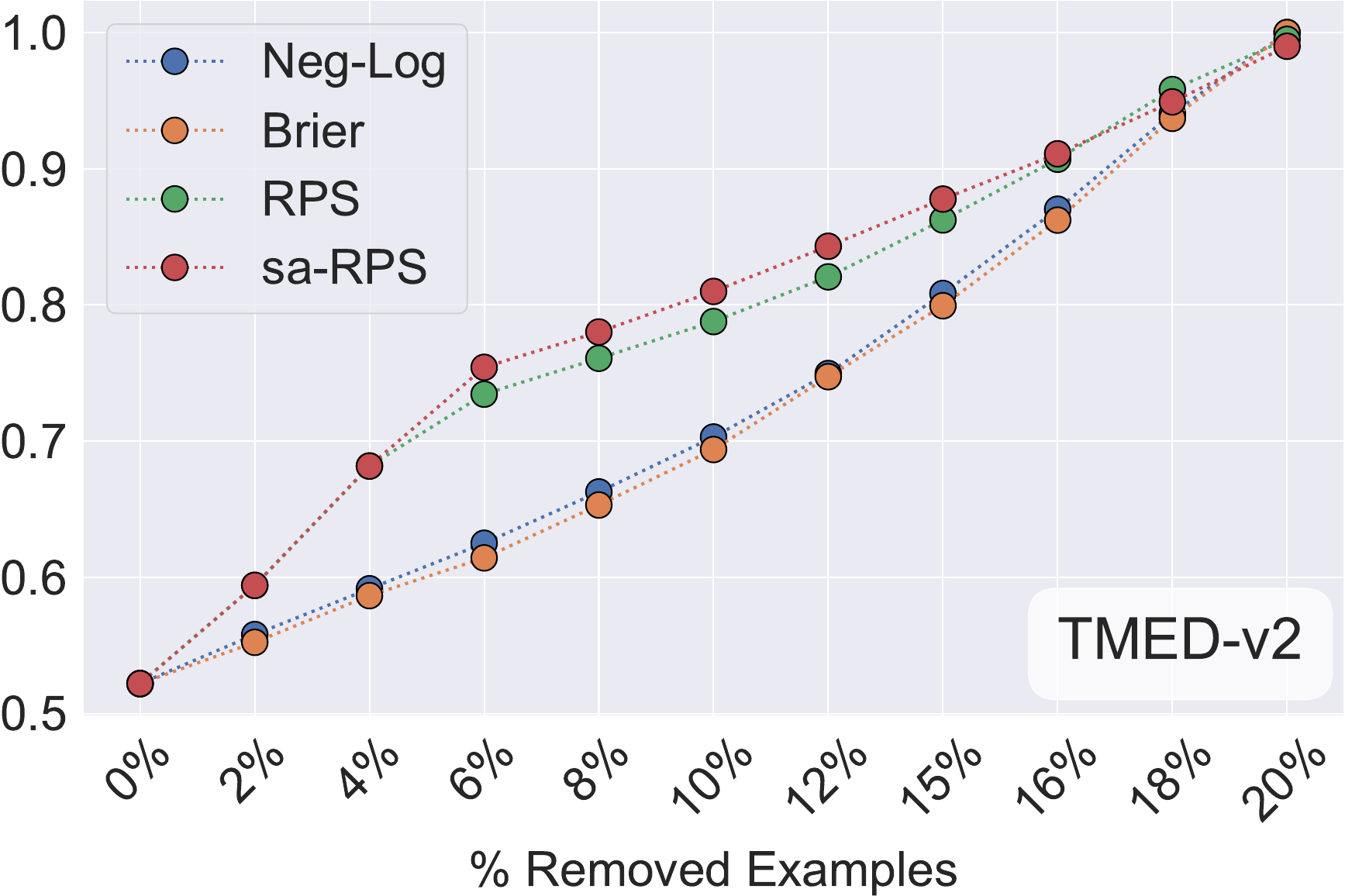}}
    \hfil
    \subfloat{\includegraphics[width = 0.495\textwidth,valign=c]{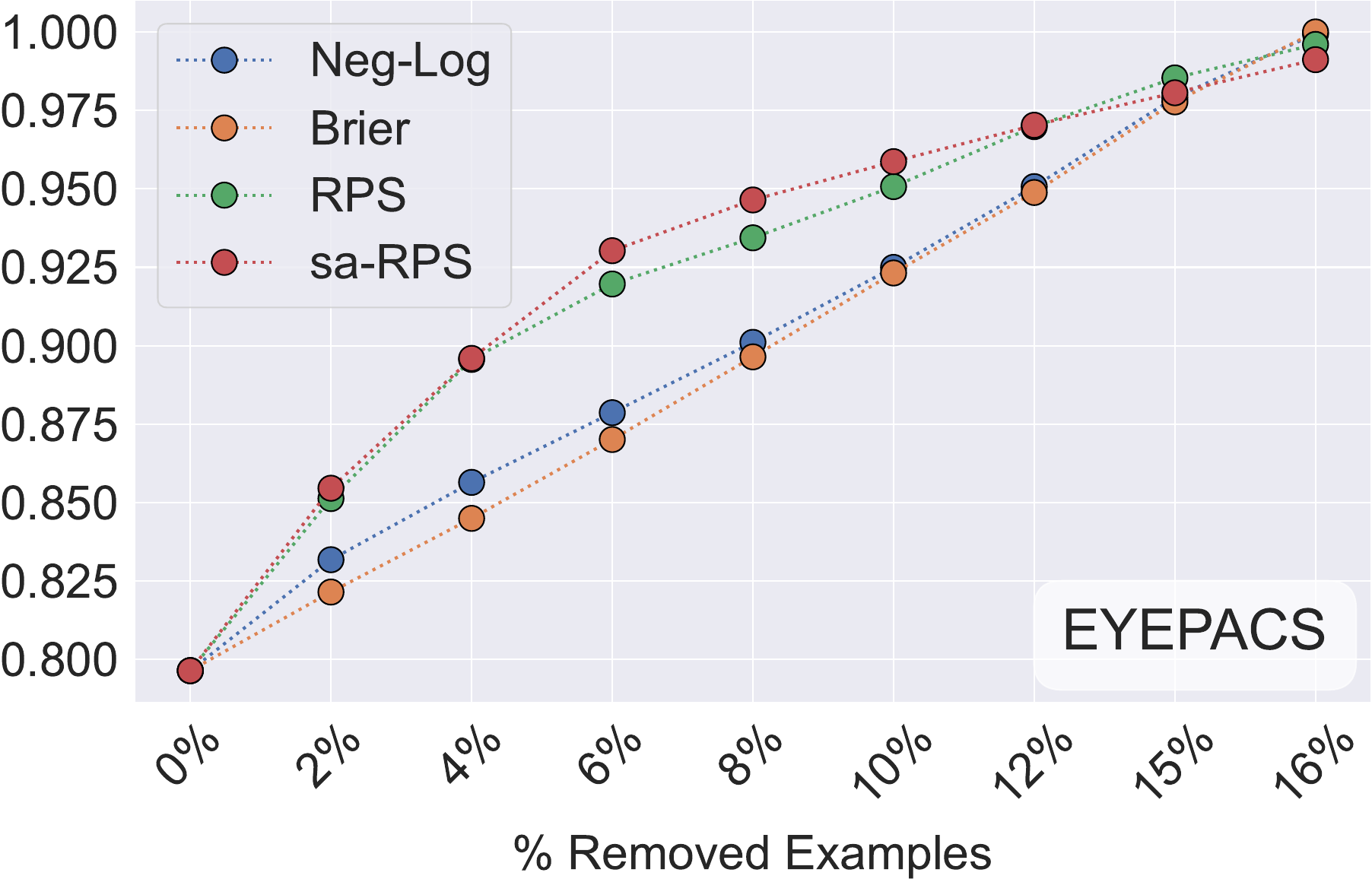}}
\caption{We sort probabilistic predictions in each test set using several PSRs: Brier, Neg-Log, RPS, sa-RPS. 
We progressively discard worse-scored samples, improving the metric of interest (only QWK shown). 
Removing worse samples according to RPS and sa-RPS leads to better QWK, implying that they both capture better ordinal classification performance at the probabilistic level.}
\label{kappa_evolution}
\end{figure}

For a visual analysis, Fig. \ref{kappa_evolution} shows the full Sample Retention Curves from which AURSC-QWK values in Table \ref{aursc_cvx} were computed.
These curves show how PSRs can indeed take a single probabilistic prediction and return a score that is correlated to QWK, which is computed over sets of samples. 
This is because as we remove samples according to any PSR, performance in the remaining test set improves in all cases.
The curves in Fig. \ref{kappa_evolution} also tell a more complete story of how the two distance-sensitive scores outperform the Brier and Neg-Log scores, particularly for TMED and Eyepacs. 
Just by removing a 5\%-6\% of samples with worse (higher) RPS, we manage to improve QWL and EC to a greater extent.

\section{Conclusion and Future Work}
We have shown that Proper Scoring Rules are useful tools for diagnosing probabilistic predictions, but the standard Brier and Logarithmic scores should not be preferred in ordinal classification problems like medical image grading. 
Instead, the Ranked Probability Score, popular in the forecasting community, should be favoured. 
We have also proposed sa-RPS, an extension of the RPS that can better handle some pathological cases.
Future work will involve using the RPS to learn ordinal classifiers, and investigating its impact in calibration problems.

\section*{Acknowledgments}
This work was supported by a Marie Skłodowska-Curie Fellowship (No 892297).

\bibliographystyle{splncs04}
\bibliography{miccai_prob_ordinal.bib}

\newpage

\begin{subappendices}
\renewcommand{\thesection}{\Alph{section}}%
\section{Further Experimental Results}\label{more_results}
In the main paper we reported results without dispersion measures due space constraints. 
Below we show expanded tables with standard deviation, and we include several more architectures (plus ConvNeXt in Tab. \ref{aursc_cvx_values}, here also with dispersion measures). 
Specifically, we include results obtained wtih a Swin-Transformer (Tab. \ref{aursc_swin}), a Densnet121 (Tab. \ref{aursc_densenet}), a Resnet50 (Tab. \ref{aursc_r50}), a Resnet18 (Tab. \ref{aursc_r18}, and a Mobilent V2 (Tab. \ref{aursc_mv2}).
In general, we observe similar trends as in the main paper. 
For all six considered neural networks, distance-sensitive PSRs always achieve greater performance than the Brier and the Logarithmic scores, and also the Square-Absolute Ranked Probability Score always outperforms the conventional Ranked Probability Score. 
It is worth stressing that when observing bootstrapped performance intervals, the neither the Brier nor the Logarithmic scores manage to overlap the SA-RPS interval in any of the two datasets, and in the Eyepacs dataset not even the best RPS result reaches the performance of worst SA-RPS result.
\vspace{-0.75cm}

\begin{table}[!h]
\renewcommand{\arraystretch}{1.00}
\setlength\tabcolsep{3.50pt}
\begin{center}
\begin{tabular}{c cc cc}
& \multicolumn{2}{c}{\textbf{TMED}} &  \multicolumn{2}{c}{\textbf{Eyepacs}} \\
\cmidrule(lr){2-3} \cmidrule(lr){4-5} 
&  {\footnotesize\textbf{AURSC-QWK}$\uparrow$}  &  {\footnotesize\textbf{AURSC-EC}$\downarrow$}   &  {\footnotesize\textbf{AURSC-QWK}$\uparrow$}  &  {\footnotesize\textbf{AURSC-EC}$\downarrow$} \\
\midrule
\textbf{Brier}       & 13.46 $\pm$ 0.35      & 3.76 $\pm$ 0.21      & 17.36 $\pm$ 0.04      & 2.84 $\pm$ 0.07       \\
\midrule
\textbf{Neg-Log}     & 13.56 $\pm$ 0.35      & 3.62 $\pm$ 0.2       & 17.44 $\pm$ 0.04      & 2.67 $\pm$ 0.07       \\
\midrule
\textbf{RPS  }       & \sd{14.76 $\pm$ 0.28} & \sd{2.68 $\pm$ 0.14} & \sd{17.81 $\pm$ 0.03} & \sd{1.99 $\pm$ 0.04}       \\
\midrule
\textbf{sa-RPS}      & \ft{14.95 $\pm$ 0.25} & \ft{2.53 $\pm$ 0.12} & \ft{17.86 $\pm$ 0.03} & \ft{1.88$\pm$ 0.04}      \\
\bottomrule
\\[-0.25cm] 
\end{tabular}
\caption{Areas under the Retained Samples Curve for \textbf{TMED} and \textbf{Eyepacs}, with a \textbf{ConvNeXt}, for each PSR; 
\unl{\textbf{best}} and \textbf{second best} values are marked.}\label{aursc_cvx_values}
\end{center}
\vspace{-2cm}
\end{table}

\begin{table}[!h]
\renewcommand{\arraystretch}{1.00}
\setlength\tabcolsep{3.50pt}
\begin{center}
\begin{tabular}{c cc cc}
& \multicolumn{2}{c}{\textbf{TMED}} &  \multicolumn{2}{c}{\textbf{Eyepacs}} \\
\cmidrule(lr){2-3} \cmidrule(lr){4-5} 
&  {\footnotesize\textbf{AURSC-QWK}$\uparrow$}  &  {\footnotesize\textbf{AURSC-EC}$\downarrow$}   &  {\footnotesize\textbf{AURSC-QWK}$\uparrow$}  &  {\footnotesize\textbf{AURSC-EC}$\downarrow$} \\
\midrule
\textbf{Brier}       & 13.60 $\pm$ 0.39      & 3.56 $\pm$ 0.25      & 17.33 $\pm$ 0.04      & 2.84 $\pm$ 0.07       \\
\midrule
\textbf{Neg-Log}       & 13.71 $\pm$ 0.39      & 3.43 $\pm$ 0.25      & 17.40 $\pm$ 0.04      & 2.68 $\pm$ 0.06       \\
\midrule
\textbf{RPS  }       & \sd{14.59 $\pm$ 0.30}  & \sd{2.75 $\pm$ 0.18} & \sd{17.77 $\pm$ 0.03} & \sd{2.02 $\pm$ 0.05}       \\
\midrule
\textbf{sa-RPS}      & \ft{14.79 $\pm$ 0.26} & \ft{2.58 $\pm$ 0.15} & \ft{17.82 $\pm$ 0.03} & \ft{1.91 $\pm$ 0.04}       \\
\bottomrule
\\[-0.25cm] 
\end{tabular}
\caption{Areas under the Retained Samples Curve for \textbf{TMED} and \textbf{Eyepacs}, with a \textbf{Swin-T}, for each PSR; 
\unl{\textbf{best}} and \textbf{second best} values are marked.}\label{aursc_swin}
\end{center}
\vspace{-2cm}
\end{table}

\begin{table}[!h]
\renewcommand{\arraystretch}{1.00}
\setlength\tabcolsep{3.50pt}
\begin{center}
\begin{tabular}{c cc cc}
& \multicolumn{2}{c}{\textbf{TMED}} &  \multicolumn{2}{c}{\textbf{Eyepacs}} \\
\cmidrule(lr){2-3} \cmidrule(lr){4-5} 
&  {\footnotesize\textbf{AURSC-QWK}$\uparrow$}  &  {\footnotesize\textbf{AURSC-EC}$\downarrow$}   &  {\footnotesize\textbf{AURSC-QWK}$\uparrow$}  &  {\footnotesize\textbf{AURSC-EC}$\downarrow$} \\
\midrule
\textbf{Brier}       & 13.25 $\pm$ 0.35      & 3.76 $\pm$ 0.2      & 17.16 $\pm$ 0.05      & 3.07 $\pm$ 0.08       \\
\midrule
\textbf{Neg-Log}     & 13.31 $\pm$ 0.34      & 3.66 $\pm$ 0.2      & 17.23 $\pm$ 0.05      & 2.91 $\pm$ 0.07       \\
\midrule
\textbf{RPS  }       & \sd{14.43 $\pm$ 0.28} & \sd{2.81 $\pm$ 0.15}& \sd{17.67 $\pm$ 0.03} & \sd{2.15 $\pm$ 0.05}       \\
\midrule
\textbf{sa-RPS}      & \ft{14.70 $\pm$ 0.23}  & \ft{2.59 $\pm$ 0.11}& \ft{17.72 $\pm$ 0.03} & \ft{2.02$\pm$ 0.04}  \\
\bottomrule
\\[-0.25cm] 
\end{tabular}
\caption{Areas under the Retained Samples Curve for \textbf{TMED} and \textbf{Eyepacs}, with a \textbf{DenseNet}, for each PSR; 
\unl{\textbf{best}} and \textbf{second best} values are marked.}\label{aursc_densenet}
\end{center}
\vspace{-2cm}
\end{table}

\begin{table}[!h]
\renewcommand{\arraystretch}{1.00}
\setlength\tabcolsep{3.50pt}
\begin{center}
\begin{tabular}{c cc cc}
& \multicolumn{2}{c}{\textbf{TMED}} &  \multicolumn{2}{c}{\textbf{Eyepacs}} \\
\cmidrule(lr){2-3} \cmidrule(lr){4-5} 
&  {\footnotesize\textbf{AURSC-QWK}$\uparrow$}  &  {\footnotesize\textbf{AURSC-EC}$\downarrow$}   &  {\footnotesize\textbf{AURSC-QWK}$\uparrow$}  &  {\footnotesize\textbf{AURSC-EC}$\downarrow$} \\
\midrule
\textbf{Brier}       & 13.14 $\pm$ 0.44      & 3.93 $\pm$ 0.27      & 16.91 $\pm$ 0.05      & 3.5 $\pm$ 0.08       \\
\midrule
\textbf{Neg-Log}       & 13.24 $\pm$ 0.43      & 3.80 $\pm$ 0.26      & 16.97 $\pm$ 0.05      & 3.34 $\pm$ 0.07       \\
\midrule
\textbf{RPS  }       & \sd{14.32 $\pm$ 0.34}      & \sd{2.98 $\pm$ 0.19}      & \sd{17.50 $\pm$ 0.03}      & \sd{2.41 $\pm$ 0.04}       \\
\midrule
\textbf{sa-RPS}       & \ft{14.46 $\pm$ 0.30}      & \ft{2.84 $\pm$ 0.16}      & \ft{17.54 $\pm$ 0.03}      & \ft{2.30 $\pm$ 0.04}   \\
\bottomrule
\\[-0.25cm] 
\end{tabular}
\caption{Areas under the Retained Samples Curve for \textbf{TMED} and \textbf{Eyepacs}, with a \textbf{ResNet50}, for each PSR; 
\unl{\textbf{best}} and \textbf{second best} values are marked.}\label{aursc_r50}
\end{center}
\vspace{-1cm}
\end{table}

\begin{table}[!h]
\renewcommand{\arraystretch}{1.03}
\setlength\tabcolsep{3.50pt}
\begin{center}
\begin{tabular}{c cc cc}
& \multicolumn{2}{c}{\textbf{TMED}} &  \multicolumn{2}{c}{\textbf{Eyepacs}} \\
\cmidrule(lr){2-3} \cmidrule(lr){4-5} 
&  {\footnotesize\textbf{AURSC-QWK}$\uparrow$}  &  {\footnotesize\textbf{AURSC-EC}$\downarrow$}   &  {\footnotesize\textbf{AURSC-QWK}$\uparrow$}  &  {\footnotesize\textbf{AURSC-EC}$\downarrow$} \\
\midrule
\textbf{Brier}       & 13.41 $\pm$ 0.41      & 3.61 $\pm$ 0.23      & 16.87 $\pm$ 0.05      & 3.59 $\pm$ 0.08       \\
\midrule
\textbf{Neg-Log}     & 13.47 $\pm$ 0.41      & 3.51 $\pm$ 0.22      & 16.93 $\pm$ 0.05      & 3.44 $\pm$ 0.08       \\
\midrule
\textbf{RPS  }       & \sd{14.49 $\pm$ 0.33}      & \sd{2.73 $\pm$ 0.16}      & \sd{17.48 $\pm$ 0.04}      & \sd{2.46 $\pm$ 0.05}       \\
\midrule
\textbf{sa-RPS}      & \ft{14.82 $\pm$ 0.29}      & \ft{2.47 $\pm$ 0.13}      & \ft{17.54 $\pm$ 0.03}      & \ft{2.31 $\pm$ 0.05}   \\
\bottomrule
\\[-0.25cm] 
\end{tabular}
\caption{Areas under the Retained Samples Curve for \textbf{TMED} and \textbf{Eyepacs}, with a \textbf{ResNet34}, for each PSR; 
\unl{\textbf{best}} and \textbf{second best} values are marked.}\label{aursc_r34}
\end{center}
\vspace{-1cm}
\end{table}

\begin{table}[!h]
\renewcommand{\arraystretch}{1.03}
\setlength\tabcolsep{3.50pt}
\begin{center}
\begin{tabular}{c cc cc}
& \multicolumn{2}{c}{\textbf{TMED}} &  \multicolumn{2}{c}{\textbf{Eyepacs}} \\
\cmidrule(lr){2-3} \cmidrule(lr){4-5} 
&  {\footnotesize\textbf{AURSC-QWK}$\uparrow$}  &  {\footnotesize\textbf{AURSC-EC}$\downarrow$}   &  {\footnotesize\textbf{AURSC-QWK}$\uparrow$}  &  {\footnotesize\textbf{AURSC-EC}$\downarrow$} \\
\midrule
\textbf{Brier}       & 13.12 $\pm$ 0.33      & 4.33 $\pm$ 0.23      & 16.74 $\pm$ 0.06      & 3.66 $\pm$ 0.08       \\
\midrule
\textbf{Neg-Log}     & 13.14 $\pm$ 0.34      & 4.25 $\pm$ 0.24      & 16.83 $\pm$ 0.05      & 3.45 $\pm$ 0.08       \\
\midrule
\textbf{RPS  }       & \sd{14.41 $\pm$ 0.27}      & \sd{3.21 $\pm$ 0.17}      & \sd{17.36 $\pm$ 0.04}      & \sd{2.55 $\pm$ 0.05}       \\
\midrule
\textbf{sa-RPS}      & \ft{14.73 $\pm$ 0.23}      & \ft{2.93 $\pm$ 0.14}      & \ft{17.41 $\pm$ 0.04}      & \ft{2.42 $\pm$ 0.05}       \\
\bottomrule
\\[-0.25cm] 
\end{tabular}
\caption{Areas under the Retained Samples Curve for \textbf{TMED} and \textbf{Eyepacs}, with a \textbf{ResNet18}, for each PSR; 
\unl{\textbf{best}} and \textbf{second best} values are marked.}\label{aursc_r18}
\end{center}
\vspace{-1cm}
\end{table}

\begin{table}[!h]
\renewcommand{\arraystretch}{1.03}
\setlength\tabcolsep{3.50pt}
\begin{center}
\begin{tabular}{c cc cc}
& \multicolumn{2}{c}{\textbf{TMED}} &  \multicolumn{2}{c}{\textbf{Eyepacs}} \\
\cmidrule(lr){2-3} \cmidrule(lr){4-5} 
&  {\footnotesize\textbf{AURSC-QWK}$\uparrow$}  &  {\footnotesize\textbf{AURSC-EC}$\downarrow$}   &  {\footnotesize\textbf{AURSC-QWK}$\uparrow$}  &  {\footnotesize\textbf{AURSC-EC}$\downarrow$} \\
\midrule
\textbf{Brier}       & 11.13 $\pm$ 0.43      & 4.92 $\pm$ 0.23      & 16.62 $\pm$ 0.05      & 3.85 $\pm$ 0.07       \\
\midrule
\textbf{Neg-Log}      & 11.13 $\pm$ 0.43      & 4.84 $\pm$ 0.23      & 16.71 $\pm$ 0.05      & 3.63 $\pm$ 0.07       \\
\midrule
\textbf{RPS  }       & \sd{12.61 $\pm$ 0.36}      & \sd{3.74 $\pm$ 0.17}      & \sd{17.28 $\pm$ 0.04}      & \sd{2.66 $\pm$ 0.05}       \\
\midrule
\textbf{sa-RPS}      & \ft{12.86 $\pm$ 0.33}      & \ft{3.46 $\pm$ 0.15}      & \ft{17.34 $\pm$ 0.03}      & \ft{2.52 $\pm$ 0.04} \\
\bottomrule
\\[-0.25cm] 
\end{tabular}
\caption{Areas under the Retained Samples Curve for \textbf{TMED} and \textbf{Eyepacs}, with a \textbf{Mobilenet}, for each PSR; 
\unl{\textbf{best}} and \textbf{second best} values are marked.}\label{aursc_mv2}
\end{center}
\vspace{-1cm}
\end{table}

\end{subappendices}

\end{document}